\documentclass[11pt]{article}
\usepackage{coling2020}
\usepackage{times}
\usepackage{url}
\usepackage{latexsym}
\usepackage{graphicx}
\usepackage{algorithm}
\usepackage{algorithmic}

\renewcommand{\algorithmiccomment}[1]{\bgroup\hfill//~#1\egroup}

\usepackage[export]{adjustbox}
\usepackage{caption}
\usepackage{booktabs}
\usepackage{multirow}
\usepackage{subfig}
\colingfinalcopy 

\title{TPLinker: Single-stage Joint Extraction of Entities and Relations Through Token Pair Linking}

\author{
  Yucheng Wang\textsuperscript{1}, Bowen Yu\textsuperscript{1}\thanks{{} {} Corresponding author.}, Yueyang Zhang\textsuperscript{2}\\ 
  \bf Tingwen Liu\textsuperscript{1}, Hongsong Zhu\textsuperscript{1}, Limin Sun\textsuperscript{1}\\
  \textsuperscript{1}School of Cyber Security, University of Chinese Academy of Sciences \\
  \textsuperscript{1}Institute of Information Engineering, Chinese Academy of Sciences \\
  \textsuperscript{2}Baidu, Inc. \\
  {\tt \{wangyucheng,yubowen,liutingwen,zhuhongsong,sunlimin\}@iie.ac.cn}\\
  {\tt zhangyueyang@baidu.com}
 }

\date{}

\begin{document}
\maketitle
\begin{abstract}
Extracting entities and relations from unstructured text has attracted increasing attention in recent years but remains challenging, due to the intrinsic difficulty in identifying overlapping relations with shared entities. Prior works show that joint learning can result in a noticeable performance gain.  However, they usually involve sequential interrelated steps and suffer from the problem of exposure bias. At training time, they predict with the ground truth conditions while at inference it has to make extraction from scratch. This discrepancy leads to error accumulation. To mitigate the issue, we propose in this paper a one-stage joint extraction model, namely, TPLinker, which is capable of discovering overlapping relations sharing one or both entities while immune from the exposure bias. TPLinker formulates joint extraction as a token pair linking problem and introduces a novel handshaking tagging scheme that aligns the boundary tokens of entity pairs under each relation type. Experiment results show that TPLinker performs significantly better on overlapping and multiple relation extraction, and achieves state-of-the-art performance on two public datasets\footnote{The source code of this paper can be obtained from https://github.com/131250208/TPlinker-joint-extraction}.
\end{abstract}

%
%
\blfootnote{
    %
    
    
    
    
    
    \hspace{-0.65cm}  
    This work is licensed under a Creative Commons 
    Attribution 4.0 International License.
    License details:
    \url{http://creativecommons.org/licenses/by/4.0/}.
}

\section{Introduction}
Extracting entities and relations from unstructured texts is  an essential step in automatic knowledge base construction~\cite{takanobu2019hierarchical}.
Traditional pipelined approaches first extract entity mentions and then classify the relation types between candidate entity pairs.
However, due to the complete separation of entity detection and relation classification, these models ignore the interaction and correlation between the two subtasks, being susceptible to cascading errors~\cite{li2014incremental}.

\begin{figure}[htb]
    \centering
    \includegraphics[scale=0.48]{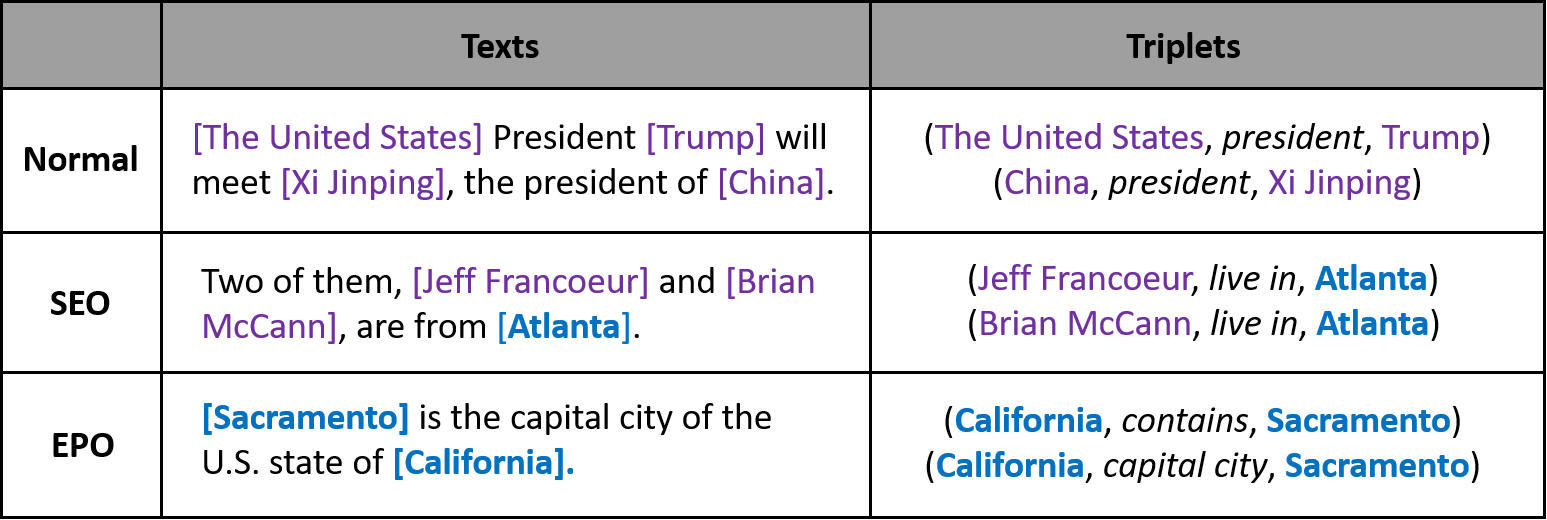}
o    \caption{Examples of the Normal, SingleEntityOverlap (SEO), and EntityPairOverlap (EPO) cases. The overlapping entities are marked in blue. }
    \label{fig:overlapping_patterns}
\vspace{-0.1in}
\end{figure}

In the last few years, there has been increasing research interest in building joint models to simultaneously extract entities and relations.
Recent works show that joint learning approaches can effectively integrate the information of entity and relation, and therefore achieve better performance in both subtasks~\cite{dai2019joint,tan2019jointly}.
~\newcite{zheng2017joint} proposed a unified tagging scheme to convert joint extraction to a sequence labeling problem but lacks the elegance to identify overlapping relations: one entity may participate in multiple relations in the same text (Figure~\ref{fig:overlapping_patterns}).

Most existing models in handling EntityPairOverlap~(EPO) and SingleEntiyOverlap~(SEO) cases can be categorized into two classes: decoder-based and decomposition-based.
Decoder-based models use encoder-decoder architecture where the decoder extracts one word or one tuple at a time like machine translation models~\cite{zeng2018extracting,nayak2019effective}.
Decomposition-based models first distinguish all the candidate subject entities that may be involved with target relations, then label corresponding object entities and relations for each extracted subject~\cite{li2019entity,yu2020jointer,wei2020CasRel}.
Although these methods have achieved reasonable performance, they all suffer from the same problem: exposure bias.
For the decoder-based method, at training time, the ground truth tokens are used as context while at inference the entire sequence is generated by the resulting model on its own, and hence the previous tokens generated by the model are fed as context.
As a result, the predicted tokens at training and inference are drawn from different distributions, namely, from the data distribution as opposed to the model distribution\cite{zhang2019bridging}.
Similarly, the decomposition-based method uses the gold subject entity as specific input to guide the model extract object entities and relations during the training process while at inference the input head-entity is given by a trained model, leading to a gap between training and inference.

In this paper, we present a one-stage method for joint extraction of entities and overlapping relations, namely TPLinker, which bridges the gap between training and inference.
TPLinker transforms the joint extraction task as a \textbf{T}oken \textbf{P}air \textbf{Link}ing problem.
Given a sentence, two positions $p_1, p_2$ and a specific relation $r$, TPLinker is to answer three Yes/No pseudo questions: ``Whether $p_1$ and $p_2$ are the start and end positions of the same entity respectively?'', ``Whether $p_1$ and $p_2$ are the start positions of two entities with $r$ relation respectively?'' and ``Whether $p_1$ and $p_2$ are the end positions of two entities with $r$ relation respectively?''
To this end, we design a handshaking tagging scheme that annotates three token link matrices for each relation to answer the above three questions.
These link matrices are then used to decode different tagging results, from which we can extract all entities and their overlapping relations. 
Intuitively, TPLinker does not contain any inter-dependency extraction steps,  so it avoids the dependence on ground truth conditions at training time, realizing the consistency of training and testing.

We evaluate our method on two public datasets: NYT~\cite{riedel2010modeling} and WebNLG~\cite{gardent2017creating}. 
Experimental results show that TPLinker outperforms previous works and achieves the state- of-the-art results on the benchmark datasets.
Further analysis demonstrates that TPLinker significantly improves the performance on Normal, SEO, EPO, and multiple relation extraction.

\section{Related Work}
Researchers proposed several methods to extract both entities and relations.
Traditional pipelined methods~\cite{zelenko2003kernel,chan2011exploiting} neglect the relevance of entity extraction and relation prediction. 
To resolve this problem, several joint models have been proposed. 
Feature-based works~\cite{yu2010jointly,miwa2014modeling} need a complicated process of feature engineering and heavily depend on NLP tools for feature extraction. 
Neural models for joint relation extraction are investigated in recent studies~\cite{gupta2016table,zheng2017joint}, they show promising results but completely giving up overlapping relations. 
To address this problem, a variety of neural networks for joint extraction of entities and overlapping relations are proposed. 
\newcite{dai2019joint} extracted triplets by tagging one sentence for $n$ times with a position-aware attention mechanism. 
\newcite{tan2019jointly} solved this task via ranking with translation mechanism. 
\newcite{takanobu2019hierarchical} firstly determined relations and then recognized entity pairs via reinforcement learning. 
\newcite{li2019entity} cast joint extraction as a multi-turn QA problem and generated questions by relation-specific templates. 
\newcite{sun2019joint} constructed an entity-relation bipartite graph to perform inference on entity types and relation types. 
\newcite{yu2020jointer} presented a unified sequence labeling framework based on a novel decomposition strategy.
However, these methods can only recognize SEO relations in the sentence and fail to extract EPO triplets.

To handle the EPO cases, \newcite{zeng2018extracting} proposed a sequence-to-sequence model to decode overlapping relations but fail to generate multi-word entities.
As the improvement, \newcite{nayak2019effective} employed an encoder-decoder model where the decoder extracts one word at a time like machine translation models.
Besides, \newcite{wei2020CasRel} proposed a novel cascade binary tagging framework that first identifies all possible subject entities in a sentence then identifies all possible relations and object entities for each subject entity.
Actually, these methods decompose the extraction of overlapping relations into several inner-dependency steps, since the decoder needs a recursive decoding process and cascade tagging has to identify subject entities in advance.
Such decomposition makes the task easy to conduct but inevitably causes the exposure bias problem, which leads to error accumulation. 
At training time, they predict the triplets with the ground truth tokens or subjects while at inference they have to rely on the predicted results.
In this paper, we propose a unified tagging method to extract entities and overlapping relations.
Different from previous methods, our model performs in one stage and generates triplets without a gap between training and inference.
\section{Methodology}

In this section, we first introduce our handshaking tagging scheme and its decoding algorithm.
Then we detail the TPLinker model structure.

\subsection{Handshaking Tagging Scheme}
%
\subsubsection{Tagging}
According to the insight that a triplet ($s$, $r$, $o$) can be determined by aligning the boundary tokens of subject entity $s$ and object entity $o$ conditioned on the relation $r$, we realize one-stage joint extraction by tagging token pairs with link labels.

\begin{figure}[t]
\centering 
\begin{minipage}[t]{0.48\textwidth}
    \includegraphics[scale=0.41]{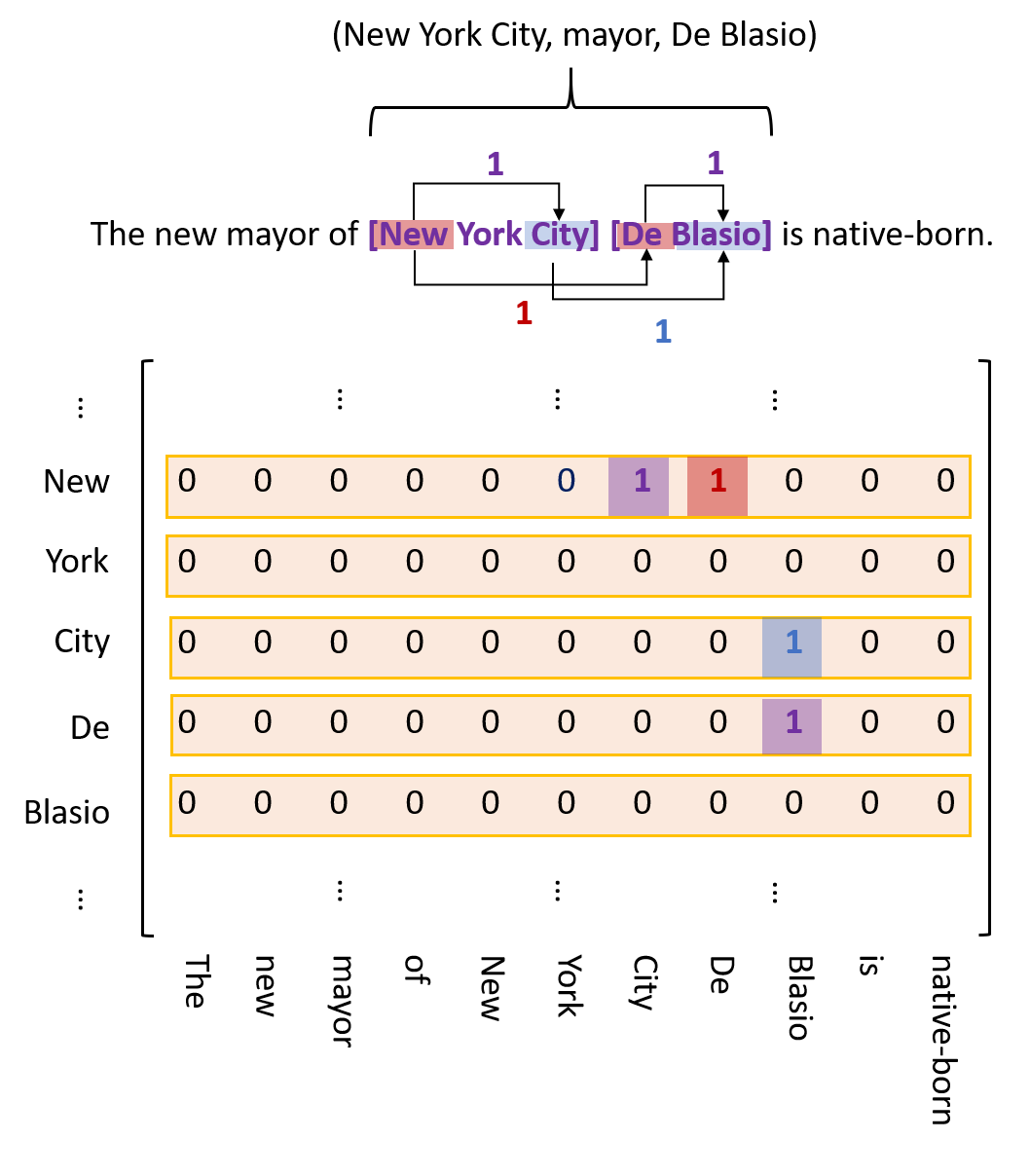}
\end{minipage}%
\begin{minipage}[t]{0.48\textwidth}
    \includegraphics[scale=0.42]{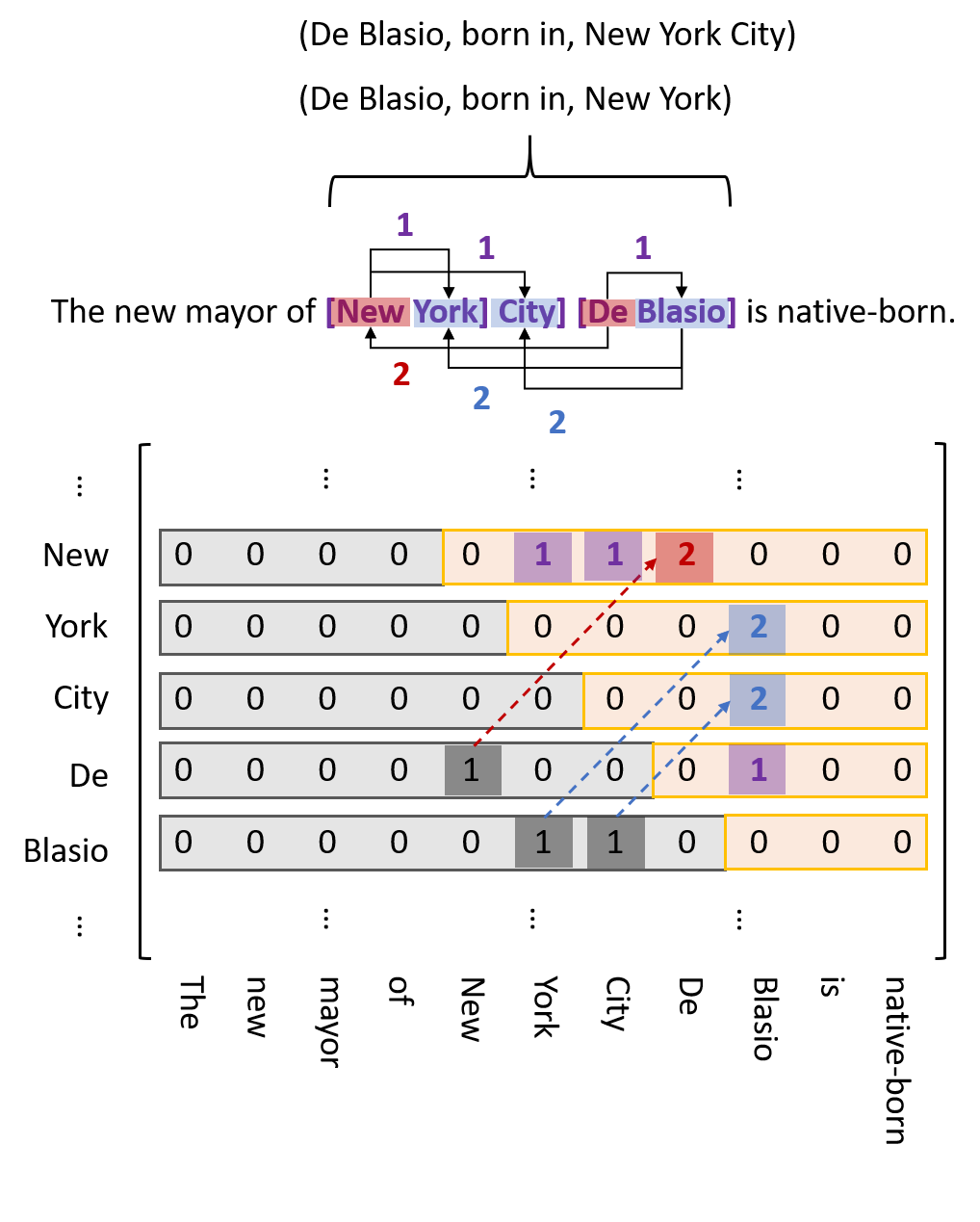}
\end{minipage}
 \caption{(Best viewed in color.) \textbf{Left}:  A tagging matrix. For the convenience to illustrate, we show all tags in one matrix, in which each color corresponds to a specific kind of tag. \textbf{Right}: An example of the Handshaking Tagging Scheme, the region in shade is not included in the tag sequence. }
 \label{fig:handshaking_tagging}
\end{figure}
As shown in the left panel of Figure~\ref{fig:handshaking_tagging}, given a sentence, we enumerate all possible token pairs and use matrices to tag token links. Formally, three types of links are defined as follows\footnote{For ease of exposition, we use "entity/subject/object head" to represent the start position of one entity/subject/object. Similarly, entity/subject/object tail is defined as the end position.}:
\textbf{1) entity head to entity tail (EH-to-ET)}. The purple tag in the matrix refers to that the two corresponding positions are respectively the start and end token of an entity. For example, ``New York City'' and ``De Blasio'' are two entities in the sentence, therefore, token pair (``New'', ``City'') and (``De'', ``Blasio'') are assigned with purple tag 1.
\textbf{2) subject head to object head  (SH-to-OH)}. The red tag means that two positions are respectively the start token of a paired subject entity and object entity. For example, there is a ``mayor'' relation between ``New York City'' and ``De Blasio'', so the token pair (``New'', and ``De'') is assigned with red tag 1.
\textbf{3) subject tail to object tail  (ST-to-OT)}. The blue tag shares a similar logic with the red tag, which means two positions are respectively the end token of a paired subject entity and object entity. For instance, the token pair (``City'', ``Blasio'') is assigned with blue tag 1.

\begin{figure}[t]
    \centering
    \includegraphics[scale=0.63]{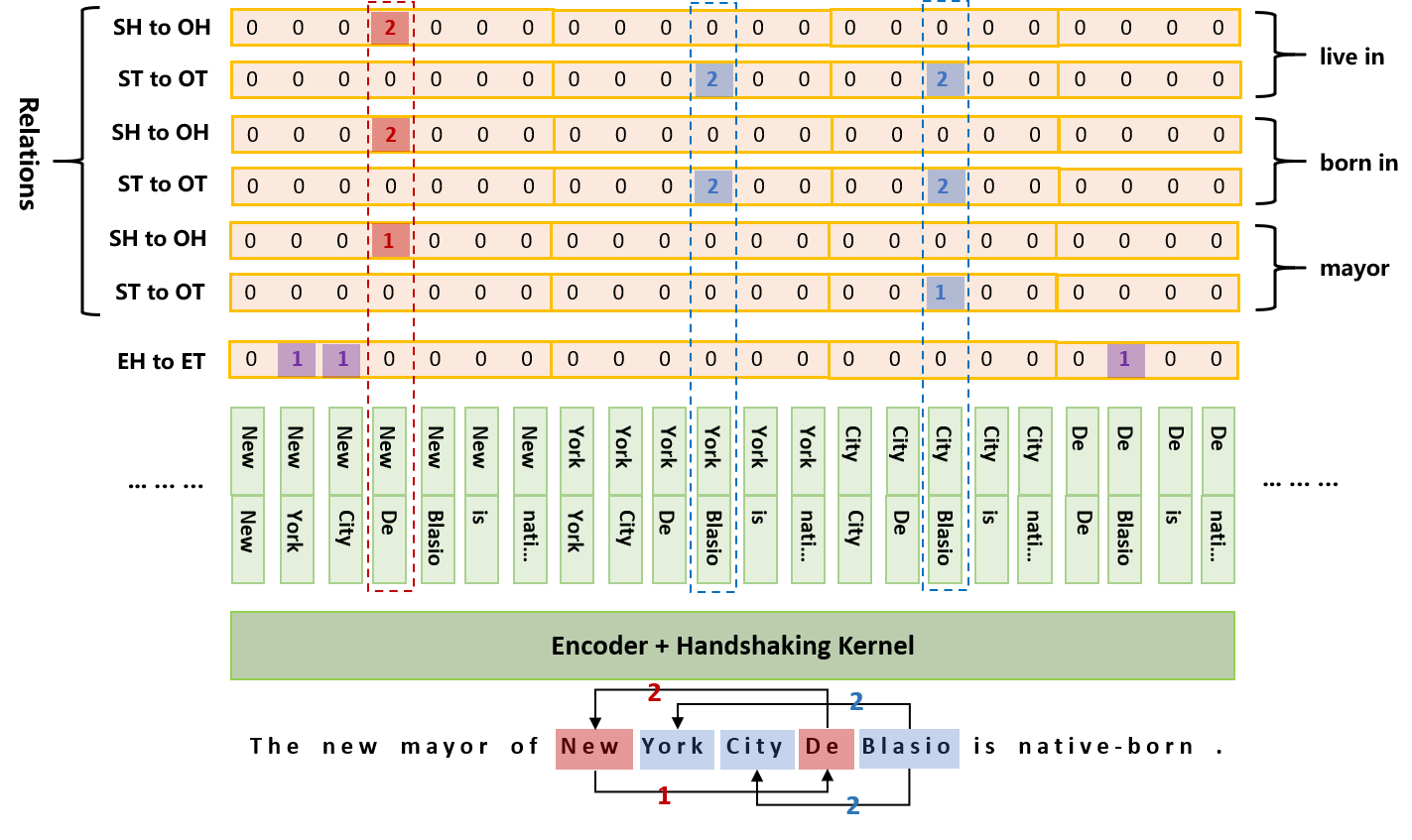}
    \caption{The Framework of TPLinker. 
    SH is short for subject head, OH is short for object head, ST is short for subject tail, and OT is short for object tail. By decoding, 5 triplets can be extracted: (New York, mayor, De Blasio), (De Blasio, born in, New York), (De Blasio, born in, New York City), (De Blasio, live in, New York), (De Blasio, live in, New York City).
}
    \label{fig:framework}
\vspace{-0.1in}
\end{figure}

As we can see from the left panel of Figure~\ref{fig:handshaking_tagging}, the matrix is quite sparse, especially the lower triangular region. Because the entity tail is impossible to appear before the entity head, the tags in the lower triangular region are all zeros, which is a huge waste of memory. 
However, the object entity could appear before the corresponding subject entity,
which means it is not reasonable to drop the lower triangular region directly. Before doing that, we map all tag 1 in the lower triangular region to tag 2 in the upper triangular region, then drop the lower triangular region.
After doing this, it is not a complete matrix anymore, in 
the practical operation, we flatten the rest items into a sequence (the orange sequences in Figure~\ref{fig:framework}) for the convenience of tensor calculation and use a map to remember the positions in the original matrix. 
The sequence is like the handshaking of all tokens, which is the reason why we refer to this scheme as the handshaking tagging scheme. 

The case in the left panel of Figure~\ref{fig:handshaking_tagging} suggests that this tagging scheme can naturally address the SingleEntiyOverlap problem and the nested entity problem from design. 
In this case, ``New York City'' and ``New York'' are nested and share the same object ``De Blasio'', which is a challenging problem for many previous methods. However, by this tagging scheme, both the three entities and the two triplets can be easily decoded (see Section~\ref{sec:decoding}). 
However, this scheme cannot handle the EntityPairOverlap problem because different relations can not be tagged together in the same matrix for the same entity pair. 
To address this issue, we do the same matrix tagging job for each relation type.
Note that EH-to-ET tagging is shared by all relations because it focuses on the general entity extraction without considering the specific relation types.
Overall, as depicted in Figure~\ref{fig:framework}, the joint extraction task is deconstructed into 2$N$+1 sequence labeling subtasks where $N$ denotes the number of pre-defined relation types, each subtask builds a tag sequence of length $\frac{n^2+n}{2}$, where $n$ is the length of the input sentence.
It seems that our tagging scheme is extremely inefficient because the length of the tagging sequence increases in a square number with increasing sentence length.
Fortunately, our experiment reveals that by utilizing a light-weight tagging model on the top of the encoder, TPLinker can achieve competitive running efficiency compared with the state-of-the-art model, since the encoder is shared by all taggers (see Figure~\ref{fig:framework}) and only needs to generate $n$ token representations for once.



\subsubsection{Decoding}\label{sec:decoding}
In the case of Figure~\ref{fig:framework}, (``New'', ``York''), (``New'', ``City'') and (``De'', ``Blasio'') are tagged as 1 in the EH-to-ET sequence, which means ``New York'', ``New York City'', and ``De Blasio'' are three entities. 
For relation ``mayor'', (``New'', ``De'') is tagged as 1 in the SH-to-OH sequence, which means the mayor of the subject starting with ``New'' is the object starting with ``De''. 
(``City'', ``Blasio'') is tagged as 1 in the ST-to-OT sequence, which means that the subject and object are the entities ending with ``City'' and ``Blasio'', respectively. 
Based on the information represented by these three sequences, a triplet can be decoded: (``New York City'', mayor, ``De Blasio''). 

The same logic goes for other relations, but note that the tag 2 has an opposite meaning to the tag 1, which represents a reversal link between tokens. 
For example, (``York'', ``Blasio'') is tagged as 2 in the ST-to-OT sequence of relation ``born in'', which means ``York'' and ``Blasio'' are respectively the tail of a paired object and subject. 
Combined with the other two sequences, the decoded triplet should be (``De Blasio'', born in, ``New York'').

Formally, the decoding process is summarized in Algorithm~\ref{alg:decoding}. 
For each relation, in the beginning, we extract all entity spans from the EH-to-ET sequence and map each head position to the corresponding entities starting with this position by a dictionary $D$. 
Next, for each relation, we firstly decode (subject tail position, object tail position) tuples from the ST-to-OT sequence and add them into a set $E$, and then decode (subject head position, object head position) tuples from the SH-to-OH sequence and lookup all possible entities starting with the head positions in the dictionary $D$.
Finally, we iterate all candidate subject-object pairs to check whether their tail positions are in  $E$. If so, a new triplet is extracted and added into the resulting set $T$. 

\begin{algorithm}[htb]
\footnotesize
\algsetup{
    linenodelimiter = {  }
}
\caption{Handshaking sequence decoding}
\label{alg:decoding}
\begin{algorithmic}[1]
\REQUIRE ~~The EH-to-ET sequence, $S_e$; \\
~~~~~~~~~The SH-to-OH sequences of relation $r$, $\{S_h^r, r\in R\}$, where $R$ is the pre-defined relation set; \\
~~~~~~~~~The ST-to-OT sequences of relation $r$, $ \{S_t^r, r \in R\}$; \\
~~~~~~~~~The map from sequence indices to matrix indices, $M$.\\
\ENSURE the predicted triplet set, $T$.
\STATE  Initialize $D \leftarrow dict$ \algorithmiccomment{the dictionary that maps entity head position to a set of entities that begin with this head position}

\STATE  Initialize $E \leftarrow set$ \COMMENT{the set of (subject tail position, object tail position)}

\STATE  Initialize $T \leftarrow set$

\FOR{$i \leftarrow 1$ \TO tag sequence length } 
\IF{$S_e[i]$ = 1}

\STATE add $M[i]$ to $D$[$M[i]$[0]] \COMMENT{$M[i]$ is a tuple (entity head position, entity tail position)}
\ENDIF
\ENDFOR

\FOR{$r \in R$} 

\FOR{$i \leftarrow 1$ \TO tag sequence length } 
\IF{$S_t^r[i]$ = 1}
\STATE add $M[i]$ to $E$ \COMMENT{$M[i]$ is a tuple (subject tail position, object tail position)}
\ELSIF{$S_t^r[i]$ = 2}
\STATE  add $(M[i][1], M[i][0])$ to $E$ \COMMENT{$M[i]$ is a tuple (object tail position, subject tail position)}
\ENDIF
\ENDFOR

\FOR{$i \leftarrow 1$ \TO tag sequence length } 
\IF{$S_h^r[i]$ = 1}
\STATE $M[i]$ is a tuple (subject head position, object head position)
\STATE $Set_{s} \leftarrow  D[M[i][0]]$ \algorithmiccomment{Set$_{s}$ records the subjects beginning with $M[i][0]$}
\STATE $Set_{o} \leftarrow  D[M[i][1]]$ \algorithmiccomment{Set$_{o}$ records the objects beginning with $M[i][0]$}
\ELSIF{$S_h^r[i]$ = 2}
\STATE $M[i]$ is a tuple (object head position, subject head position
\STATE $Set_{s} \leftarrow D[M[i][1]]$ 
\STATE $Set_{o} \leftarrow  D[M[i][0]]$
\ENDIF

\FOR{$s \in Set_{s}$}
\FOR{$o \in Set_{o}$}
\IF{$ (s[1], o[1]) \in E$}
\STATE add $(s, r, o)$ to $T$
\ENDIF
\ENDFOR
\ENDFOR
\ENDFOR
\ENDFOR

\RETURN $T$
\end{algorithmic}
\end{algorithm}
\subsection{Token Pair Representation}


Given a sentence $[w_1,\cdots,w_n]$ of length $n$, we first map each token $w_i$ into a low-dimensional contextual vector $\mathbf{h_i}$ by a basic encoder. Then we can generate a representation $\mathbf{h}_{i, j}$ for the token pair ($w_i,w_j$) as follows:
%
\begin{equation}
    \label{equ:tok_pair_rep}
    \mathbf{h}_{i, j} = \tanh(\mathbf{W}_h \cdot [\mathbf{h_i}; \mathbf{h_j}] + \mathbf{b}_h), j \geq i,
\end{equation}
where $\mathbf{W}_h$ is a parameter matrix and $\mathbf{b}_h$ is a bias vector to be learned during training.
Equation~\ref{equ:tok_pair_rep} is also denoted as Handshaking Kernel in Figure~\ref{fig:framework}.

\subsection{Handshaking Tagger}
We utilize a unified architecture for EH-to-ET, SH-to-OH and ST-to-OT tagging. 
Given a token pair representation $\mathbf{h}_{i, j}$, the link label of token pair ($w_i,w_j$) is predicted by Equation~\ref{equ:link_tag}.
\begin{equation}
    \label{equ:prob_link}
    P(y_{i,j}) = {\rm{Softmax}}(\mathbf{W}_o \cdot \mathbf{h}_{i, j} + \mathbf{b}_o),
\end{equation}
\begin{equation}
    \label{equ:link_tag}
    link(w_i, w_j) =  \mathop{\arg\max_{l}} P(y_{i,j} = l),
\end{equation}
where $P(y_{i,j} = l)$ represents the probability of identifying the link of ($w_i,w_j$) as $l$. 

\subsection{Loss Function}
We define the training loss as below:
\begin{equation}
    \label{equ:loss}
    L_{link} = -\frac{1}{N}\sum_{i = 1, j \geq i}^N ~ \sum_{* \in \{E, H, T\}} \log P(y^*_{i,j} = \hat{l}^*)
\end{equation}



Here, $N$ is the length of the input sentence, $\hat{l}$ is the true tag, $E$, $H$, and $T$ denote the taggers of EH-to-ET, SH-to-OH and ST-to-OT, respectively. 
\section{Experiments}

\subsection{Datasets}
For the convenience to compare our model with previous work, we follow the popular choice of datasets: NYT~\cite{riedel2010modeling} and WebNLG~\cite{gardent2017creating}. 
There are two versions of these two datasets according to the annotation standard: 1) annotating the last word of the entities and 2) annotating the whole entity span. 
~\newcite{zeng2018extracting}, ~\newcite{zeng2019learning}, \newcite{fu2019graphrel} and \newcite{wei2020CasRel} used the first version while \newcite{yu2020jointer} chose the second version.
For fair comparisons, we  evaluate our model on both settings. 
The first version datasets are denoted as NYT$^{\star}$ and WebNLG$^{\star}$ and the second ones are denoted as NYT and WebNLG, respectively.
Following \newcite{zeng2018extracting}, to further study the capability of TPLinker in extracting overlapping and multiple relations, we also split the test set by overlapping patterns and triplet number. The statistics are reported in Table~\ref{table:statistics}. 

\begin{table*}[t]
\small
\begin{center}
{
    \caption{Statistics of the datasets in our experiments.
    Normal set contains sentences without overlapping triple. SPO is short for Single Pair Overlapping. EPO is short for Entity Pair Overlapping. Note that the relation number of WebNLG and WebNLG$^{\star}$ were miswritten as 246 in \cite{wei2020CasRel} and \cite{yu2020jointer}, which is the number of the original WebNLG dataset instead of the subsets they used. We recount and give the correct numbers. $\star$ marks the relation number of WebNLG$^{\star}$.}
    \label{table:statistics}
} 
    \setlength{\tabcolsep}{1.5mm}{\begin{tabular}{lcccccccccccc}
    \toprule
        \multirow{2}{*}{Model} & \multirow{2}{*}{Train} 
        & \multirow{2}{*}{Valid} 
        & \multirow{2}{*}{Test} 
        & \multicolumn{3}{c}{Overlapping Pattern}
        & \multicolumn{5}{c}{Number of triplets}
        & \multirow{2}{*}{Relation} \\
        \multicolumn{4}{c}{}
         & Normal & SEO & EPO & N = 1 & N = 2 & N = 3 & N = 4 & N $\geq$ 5 & \\
    \midrule
        NYT & 56195 & 5000 & 5000 & 3266 & 1297 & 978 & 3244 & 1045 & 312 & 291 & 108 & 24\\
        WebNLG & 5019 & 500 & 703 & 246 & 457 & 26 & 266 & 171 & 131 & 90 &  45 & 216/171$^\star$ \\
    \bottomrule
    \end{tabular}}
\end{center}
\end{table*}

\subsection{Evaluation}
In our experiments, to keep in line with previous works, we use \textbf{Partial Match} for NYT$^\star$ and WebNLG$^\star$, an extracted triplet is regarded as correct if the relation and the head of both subject entity and object entity are all correct, and \textbf{Exact Match} for NYT and WebNLG, the whole spans of subject and object are needed to be matched.  
We follow popular choice to report the standard micro Precision (Prec.), Recall (Rec.), and F1-score as in line with all the baselines.

\subsection{Implementation Details}\label{sec:implementation details}
Our TPLinker is implemented with PyTorch and the network weights are optimized with Adam~\cite{kingma2014adam}.
We try two encoders in this paper.
One is the combination of 300-dimensional GloVe embeddings\cite{pennington2014glove} and 2-layer stacked BiLSTM, the hidden dimension of the 2 layers are set as 300 and 600 respectively.
Dropout is applied to word embeddings and hidden states with a rate of 0.1.
Another is BERT, where we use the base cased English model\footnote{Available at https://huggingface.co/bert-base-cased}.
The learning rate is set as 1e-3/5e-5 in the backbone of BiLSTM/BERT.
We also conduct Cosine Annealing Warm Restarts learning rate schedule~\cite{loshchilov2016sgdr}.
Following previous works~\cite{zeng2018extracting,fu2019graphrel,wei2020CasRel}, we set the max length of input sentence to 100.
The batch size is set as 24/6 in NYT/WebNLG.
We use Tesla V100 to train the model for at most 100 epochs and choose the model with the best performance on the validation set to output results on the test set. 


\subsection{Comparison Models}
For comparison, we employ the following models as baselines:
(1) \textbf{NovelTagging}\cite{zheng2017joint} applies a novel tagging strategy that incorporates both entity and relation roles, thus the joint extraction task is converted to a sequence labeling problem. This model fails to solve the overlapping problem;
(2) \textbf{CopyRE}~\cite{zeng2018extracting} first explores the encoder-decoder architecture for this task, trying to face overlapping problem by generating all triples in the sentence. This model can only copy the last word of an entity;
(3) \textbf{MultiHead}\cite{bekoulis2018joint} first identifies all candidate entities, then formulates the task as a multi-head selection problem;
(4) \textbf{GraphRel}\cite{fu2019graphrel} utilizes graph convolutional network to extract overlapping relations by splitting entity mention pairs into several word pairs and considering all pairs for prediction;
(5) \textbf{OrderCopyRE}\cite{zeng2019learning} is an extension of CopyRE, which applies the reinforcement learning into an encoder-decoder model to generate multiple triplets;
(6) \textbf{ETL-Span}\cite{yu2020jointer} applies a span-based tagging strategy and hierarchically decode triplets to model the internal dependencies;
(7) \textbf{WDec}\cite{nayak2019effective} is the improvement of CopyRE, which extracts one word at each time step, can extract overlapping relations and triplets with multi-token entities;
(8) \textbf{CasRel}\cite{wei2020CasRel} is the state-of-the-art method on the NYT and WebNLG datasets based on the BERT backbone, which first identifies all possible head-entities in a sentence then identifies all possible relations and corresponding tail-entities for each head-entity.

\subsection{Experimental Results and Analysis}
\subsubsection{Main Results}
Table~\ref{table:main_res} reports the results of our models against other baseline methods on all datasets.
We can observe that TPLinker outperforms all the baselines in terms of F1-score. 
Especially, TPLinker$_{BERT}$ improves 2.3 percentages on NYT$^\star$, 14.0 on NYT, and 3.6 on WebNLG over the state-of-the-art models. 
To validates the utility of our handshaking tagging scheme, we ablate the BERT and use BiLSTM as the substituted encoder to output results. 
It can be seen that TPLinker$_{BiLSTM}$ is still very competitive to existing state-of-the-art models, CasRel$_{BERT}$. What is more, TPLinker$_{BiLSTM}$ outperforms CasRel$_{BiLSTM}$ by 6.8 percentages on WebNLG$^\star$, which suggests the superiority of our scheme.
Even though the two SOTA models, CasRel and ETL-Span, achieve encouraging scores, they still suffer from some problems. 
For CasRel, it is inherently a two-stage method, which suffers from both exposure bias and error propagation. 
For ETL-Span, it has the same problems with CasRel and fails to handle the EPO problem. 
TPLinker solves all these problems and offers a bonus capability that it can extract a triplet with nested entities. For all we know, CasRel and ETL-Span cannot extract nested subject entities.

TPLinker performs well on both the dataset annotating the last word and the one annotating the whole span. 
Supporting this point, TPLinker achieves almost the same scores on NYT and NYT$^\star$, which also suggests that the dataset only annotating the last word is not always easier than the one that annotates the whole span. 
Even on the contrary, it could be harder because different entities may share the same last word, which makes the number of overlapping cases increase.
Granted, We figure out that there is a significant gap between the performance on WebNLG and WebNLG$^\star$, which may disprove the above point. However, we also find out that there are 127 wrong triplets in WebNLG, containing a meaningless empty entity. Plus, WebNLG has 216 relations but WebNLG$^\star$ has only 171. Therefore, we attribute the performance gap to these two reasons. For a fair comparison with the state-of-the-art model on this dataset, we do not fix these problems in the experiments.

We can also observe that TPLinker$_{BERT}$ achieves a similar F1 score with CasRel$_{BERT}$ on WebNLG$^\star$. 
We consider that it is because (1) the performances on WebNLG$^\star$ are already saturated since extracting triplets with 171 predefined relations is very hard, especially the training data is very small (5019 sentences). These methods achieving a 90+ F1 score might have already surpassed human-level performance. In other words, the room for boosting is too limited. (2) Many relations in WebNLG$^\star$ share the same meaning, e.g. \textit{LeaderName} and \textit{Leader}, \textit{affiliation} and \textit{affiliations}, which are confusing to the model. In many cases, our model will extract both of them, but normally only one of them has been annotated in the test set. The absence of these correct annotations seriously hurts the precision of TPLinker$_{BERT}$ because our model has greater ability in extracting overlapping relations and achieves better recall rate. 

\begin{table*}[t]
\small
\begin{center}
{
    \caption{Main results. Bold marks the highest score. $\ddagger$ marks results quoted directly from the original papers. $^\dagger$ marks results reported by \cite{dai2019joint} and \cite{zeng2018extracting}. $^*$ marks results produced with official implementation. $^\star$ marks the datasets that only annotating the last word. }
    \label{table:main_res}
} 
    \setlength{\tabcolsep}{1.5mm}{\begin{tabular}{lccccccccccccc}
    \toprule
        \multirow{2}{*}{Model} & \multicolumn{3}{c}{NYT$^\star$} & \multicolumn{3}{c}{NYT}& \multicolumn{3}{c}{WebNLG$^\star$} & \multicolumn{3}{c}{WebNLG}\\
         & Prec. & Rec. & F1 & Prec. & Rec. & F1 & Prec. & Rec. & F1 & Prec. & Rec. & F1 \\
    \midrule
        NovelTagging$^\dagger$ \cite{zheng2017joint} & -- & -- & -- & 32.8 & 30.6 & 31.7 & -- & -- & -- & 52.5 & 19.3 & 28.3 \\
        CopyRE$^\ddagger$ \cite{zeng2018extracting}& 61.0 & 56.6 & 58.7 & -- & -- & -- & 37.7 & 36.4 & 37.1 & -- & -- & --  \\
        MultiHead$^*$ \cite{bekoulis2018joint} & -- & -- & -- & 60.7 & 58.6 & 59.6 & -- & -- & -- & 57.5 & 54.1 & 55.7 \\
        GraphRel$^\ddagger$ \cite{fu2019graphrel} & 63.9 & 60.0 & 61.9 & -- & -- & -- & 44.7 & 41.1 & 42.9 & -- & -- & -- \\
        OrderCopyRE$^\ddagger$ \cite{zeng2019learning}& 77.9 & 67.2 & 72.1 & -- & -- & -- & 63.3 & 59.9 & 61.6 & -- & -- & --  \\
        ETL-Span$^\ddagger$$^*$\cite{yu2020jointer} & 84.9 & 72.3 & 78.1 & 85.5 & 71.7 & 78.0 & 84.0 & 91.5 & 87.6 & 84.3  & 82.0 & 83.1 \\
        WDec$^\ddagger$\cite{nayak2019effective} & \textbf{94.5} & 76.2 & 84.4 & -- & -- & -- & -- & -- & -- & -- & -- & --  \\
        CasRel$_{LSTM}^\ddagger$\cite{wei2020CasRel} & 84.2 & 83.0 & 83.6 & -- & -- & -- & 86.9 & 80.6 & 83.7 & -- & -- & --  \\
        CasRel$_{BERT}^\ddagger$\cite{wei2020CasRel} & 89.7 & 89.5 & 89.6 & -- & -- & -- & \textbf{93.4} & 90.1 & 91.8 & -- & -- & --  \\
        \midrule
        TPLinker$_{LSTM}$ & 83.8 & 83.4 & 83.6 & 86.0 & 82.0 & 84.0 & 90.8 & 90.3 & 90.5 & \textbf{91.9}  & 81.6 & 86.4 \\
        TPLinker$_{BERT}$ & 91.3 & \textbf{92.5} & \textbf{91.9} & \textbf{91.4} & \textbf{92.6} & \textbf{92.0} & 91.8 & \textbf{92.0} & \textbf{91.9} & 88.9 & \textbf{84.5} & \textbf{86.7}  \\
    \bottomrule
    \end{tabular}}
\end{center}
\end{table*}

\subsubsection{Analysis on Different Sentence Types}
To verify the ability of our model in handling the overlapping problem and extracting multiple relations, we conduct further experiments on different subsets of NYT$^{\star}$ and WebNLG$^{\star}$. 

Results suggest the advantages of TPLinker in handling sentences with overlapping relations or multiple relations. As shown in Table~\ref{table:f1_on_split}, most baselines present an obvious decreasing trend with the increasing complexity of sentences. However, TPLinker presents a significant improvement on the harder sentences, ones with overlapping relations, or more than two relations. Even though CasRel also presents an increasing trend, TPLinker outperforms CasRel on all subsets except some simpler ones without overlapping or with less than three relations. Especially, TPLinker outperforms CasRel by 6.3 percentages on sentences with more than or equal to 5 relations.


\begin{table*}[t]
\small
\begin{center}
{
    \caption{F1 score on sentences with different overlapping pattern and different triplet number. Results of baselines are all quoted directly from \cite{wei2020CasRel} except for ETL-Span, of which the results are reproduced by the official implementation.
    }
    \label{table:f1_on_split}
} 
    \setlength{\tabcolsep}{0.5 mm}{\begin{tabular}{lccccccccccccccccc}
    \toprule
        \multirow{2}{*}{Model} & \multicolumn{8}{c}{NYT$^\star$}& \multicolumn{9}{c}{WebNLG$^\star$} \\
         & Normal & SEO & EPO & N = 1 & N = 2 & N = 3 & N = 4 & N $\geq$ 5 && Normal & SEO & EPO & N = 1 & N = 2 & N = 3 & N = 4 & N $\geq$ 5 \\
    \midrule
        CopyRE & 66.0 & 48.6 & 55 & 67.1 & 58.6 & 52.0 & 53.6 & 30.0 && 59.2 & 33.0 & 36.6 & 59.2 & 42.5 & 31.7 & 24.2 & 30.0 \\
        GraphRel & 69.6 & 51.2 & 58.2 & 71.0 & 61.5 & 57.4 & 55.1 & 41.1 && 65.8 & 38.3 & 40.6 & 66.0 & 48.3 & 37.0 & 32.1 & 32.1  \\
        OrderCopyRE & 71.2 & 69.4 & 72.8 & 71.7 & 72.6 & 72.5 & 77.9 & 45.9 && 65.4 & 60.1 & 67.4 & 63.4 & 62.2 & 64.4 & 57.2 & 55.7 \\
        ETL-Span & 88.5  & 87.6 & 60.3 & 85.5 & 82.1 & 74.7  & 75.6 & 76.9  && 87.3 & 91.5 & 80.5 & 82.1 & 86.5 & 91.4 & 89.5 & 91.1 \\
        CasRel$_{BERT}$ & 87.3 & 91.4 & 92 & 88.2 & 90.3 & 91.9 & 94.2 & 83.7 &&  \textbf{89.4} & 92.2\footnotemark[5] & 94.7\footnotemark[5] & \textbf{89.3} & \textbf{90.8} & 94.2 & 92.4 & 90.9  \\
        \midrule
        TPLinker$_{LSTM}$ & 80.7 & 85.5 & 86.5 & 80.8 & 85.4 & 85.8 & 87.8 & 81.5 && 86.2 & 91.4 & 92.5 & 85.9 & 89.1 & 92.9 & 92.4 & 91.4 \\
        TPLinker$_{BERT}$ & \textbf{90.1} & \textbf{93.4} & \textbf{94.0} & \textbf{90.0} & \textbf{92.8} & \textbf{93.1} & \textbf{96.1} & \textbf{90.0} && 87.9 & \textbf{92.5} & \textbf{95.3} & 88.0 & 90.1 & \textbf{94.6} & \textbf{93.3} & \textbf{91.6} \\
        
    \bottomrule
    \end{tabular}}
\end{center}
\end{table*}
                                
\footnotetext{The original paper confuses the score on WebNLG-EPO and the score on WebNLG-SEO. The author has corrected them and updated the paper on arXiv (https://arxiv.org/abs/1909.03227). We use the correct scores here.}

\subsubsection{Analysis on Computational Efficiency}


\begin{table*}[t]
\small
\begin{center}
{
    \caption{Comparison on Computational Efficiency. Params$_{all}$ denotes the number of parameters of the entire model. Prop$_{encoder}$ is the proportion of encoder parameters in the total model parameters. Inference Time represents the 
    the average time (ms) the model takes to process a sample. $\dagger$ marks the inference time when the batch size is set to 1. }
    \label{table:statistics of costs}
} 
    \setlength{\tabcolsep}{1.5mm}{\begin{tabular}{lccccccccc}
    \toprule
        \multirow{2}{*}{Model} 
        & \multicolumn{3}{c}{NYT$^\star$}
        & \multicolumn{3}{c}{WebNLG$^\star$}\\
        & Params$_{all}$ & Prop$_{encoder}$ & Inference Time & Params$_{all}$ & Prop$_{encoder}$ & Inference Time \\
    \midrule
        CasRel$_{BERT}$ & 107,719,680
        & 99.96\% & 54.0 & 107,984,216 & 99.76\%  & 76.8 \\
        TPLinker$_{BERT}$ & 109,602,962 & 98.82\%  & 15.2 / 82.7$^\dagger$ & 110,281,220 & 98.21\% & 25.6 / 112.6$^\dagger$ \\
    \bottomrule
    \end{tabular}}
\end{center}
\end{table*}

Table~\ref{table:statistics of costs} shows the comparison of computational efficiency between CasRel$_{BERT}$ and TPLinker$_{BERT}$. In this comparison experiment, we use the official implementation and default configuration of CasRel$_{BERT}$ to produce the statistics data. In the inference phase, the decoding speed of TPLinker$_{BERT}$ is almost 3.6 times as CasRel$_{BERT}$. Since the number of subjects is uncertain in a given sentence, it is difficult for CasRel$_{BERT}$ to predict objects in batch for each subject. In the official implementation, CasRel$_{BERT}$ is restricted to processes one sentence at a time, which means it is seriously inefficient and difficult to deploy. 
On the contrary, TPLinker$_{BERT}$ is capable of handling data in batch mode because it is a one-stage model. 
Besides, even though we set the batch size of TPLinker$_{BERT}$ to 1, the inference speed is still competitive, which again confirms the efficiency of TPLinker$_{BERT}$. Actually, CasRel$_{BERT}$ and TPLinker$_{BERT}$ both use BERT as the basic encoder, which is the most time-consuming part and takes up the most of model parameters, so the time cost of handshaking tagging is not significant.


\section{Conclusion}
In this paper, we propose an end-to-end sequence labeling model TPLinker for joint extraction of entities and relations based on a novel handshaking tagging strategy, by which the joint extraction task is converted to a token pair linking game. 
To the best of our knowledge, TPLinker is the first one-stage joint extraction model that can extract all kinds of overlapping relations without the influence of exposure bias.
Experimental results show that our model outperforms all baselines and achieves a new state-of-the-art on two public datasets. 
Further analysis especially demonstrates the capabilities of our model on handling sentences with overlapping relations and multiple relations. 
The results also prove that it is of benefit to close up the gap between training and inference. 
In the future, we would like to generalize the token linking idea and explore its performance on other information extraction problems, such as nested name entity extraction and event extraction. 

\section*{Acknowledgements}
This work was supported by the National Key R\&D Program of China (Grant No. 2017YFB0802804).


\bibliographystyle{coling}
\bibliography{coling2020}

\begin{thebibliography}{}

\bibitem[\protect\citename{Bekoulis \bgroup et al.\egroup
  }2018]{bekoulis2018joint}
Giannis Bekoulis, Johannes Deleu, Thomas Demeester, and Chris Develder.
\newblock 2018.
\newblock Joint entity recognition and relation extraction as a multi-head
  selection problem.
\newblock {\em Expert Systems with Applications}.

\bibitem[\protect\citename{Chan and Roth}2011]{chan2011exploiting}
Yee~Seng Chan and Dan Roth.
\newblock 2011.
\newblock Exploiting syntactico-semantic structures for relation extraction.
\newblock In {\em Proceedings of ACL}.

\bibitem[\protect\citename{Dai \bgroup et al.\egroup }2019]{dai2019joint}
Dai Dai, Xinyan Xiao, Yajuan Lyu, Shan Dou, Qiaoqiao She, and Haifeng Wang.
\newblock 2019.
\newblock Joint extraction of entities and overlapping relations using
  position-attentive sequence labeling.
\newblock In {\em Proceedings of AAAI}.

\bibitem[\protect\citename{Fu \bgroup et al.\egroup }2019]{fu2019graphrel}
Tsu-Jui Fu, Peng-Hsuan Li, and Wei-Yun Ma.
\newblock 2019.
\newblock Graphrel: Modeling text as relational graphs for joint entity and
  relation extraction.
\newblock In {\em Proceedings of ACL}.

\bibitem[\protect\citename{Gardent \bgroup et al.\egroup
  }2017]{gardent2017creating}
Claire Gardent, Anastasia Shimorina, Shashi Narayan, and Laura
  Perez-Beltrachini.
\newblock 2017.
\newblock Creating training corpora for nlg micro-planning.
\newblock In {\em Proceedings of ACL}.

\bibitem[\protect\citename{Gupta \bgroup et al.\egroup }2016]{gupta2016table}
Pankaj Gupta, Hinrich Schtze, and Bernt Andrassy.
\newblock 2016.
\newblock Table filling multi-task recurrent neural network for joint entity
  and relation extraction.
\newblock In {\em Proceedings of COLING}.

\bibitem[\protect\citename{Kingma and Ba}2014]{kingma2014adam}
Diederik~P Kingma and Jimmy Ba.
\newblock 2014.
\newblock Adam: A method for stochastic optimization.
\newblock {\em arXiv preprint arXiv:1412.6980}.

\bibitem[\protect\citename{Li and Ji}2014]{li2014incremental}
Qi~Li and Heng Ji.
\newblock 2014.
\newblock Incremental joint extraction of entity mentions and relations.
\newblock In {\em Proceedings of ACL}.

\bibitem[\protect\citename{Li \bgroup et al.\egroup }2019]{li2019entity}
Xiaoya Li, Fan Yin, Zijun Sun, Xiayu Li, Arianna Yuan, Duo Chai, Mingxin Zhou,
  and Jiwei Li.
\newblock 2019.
\newblock Entity-relation extraction as multi-turn question answering.
\newblock In {\em Proceedings of ACL}.

\bibitem[\protect\citename{Loshchilov and Hutter}2016]{loshchilov2016sgdr}
Ilya Loshchilov and Frank Hutter.
\newblock 2016.
\newblock Sgdr: Stochastic gradient descent with warm restarts.
\newblock {\em arXiv preprint arXiv:1608.03983}.

\bibitem[\protect\citename{Miwa and Sasaki}2014]{miwa2014modeling}
Makoto Miwa and Yutaka Sasaki.
\newblock 2014.
\newblock Modeling joint entity and relation extraction with table
  representation.
\newblock In {\em Proceedings of EMNLP}, pages 1858--1869.

\bibitem[\protect\citename{Nayak and Ng}2020]{nayak2019effective}
Tapas Nayak and Hwee~Tou Ng.
\newblock 2020.
\newblock Effective modeling of encoder-decoder architecture for joint entity
  and relation extraction.
\newblock In {\em Proceedings of AAAI}.

\bibitem[\protect\citename{Pennington \bgroup et al.\egroup
  }2014]{pennington2014glove}
Jeffrey Pennington, Richard Socher, and Christopher~D Manning.
\newblock 2014.
\newblock Glove: Global vectors for word representation.
\newblock In {\em Proceedings of EMNLP}.

\bibitem[\protect\citename{Riedel \bgroup et al.\egroup
  }2010]{riedel2010modeling}
Sebastian Riedel, Limin Yao, and Andrew McCallum.
\newblock 2010.
\newblock Modeling relations and their mentions without labeled text.
\newblock In {\em ECML-PKDD}.

\bibitem[\protect\citename{Sun \bgroup et al.\egroup }2019]{sun2019joint}
Changzhi Sun, Yeyun Gong, Yuanbin Wu, Ming Gong, Daxin Jiang, Man Lan, and Sun.
\newblock 2019.
\newblock Joint type inference on entities and relations via graph
  convolutional networks.
\newblock In {\em Proceedings of ACL}.

\bibitem[\protect\citename{Takanobu \bgroup et al.\egroup
  }2019]{takanobu2019hierarchical}
Ryuichi Takanobu, Tianyang Zhang, Jiexi Liu, and Minlie Huang.
\newblock 2019.
\newblock A hierarchical framework for relation extraction with reinforcement
  learning.
\newblock In {\em Proceedings of AAAI}.

\bibitem[\protect\citename{Tan \bgroup et al.\egroup }2019]{tan2019jointly}
Zhen Tan, Xiang Zhao, Wei Wang, and Weidong Xiao.
\newblock 2019.
\newblock Jointly extracting multiple triplets with multilayer translation
  constraints.
\newblock In {\em Proceedings of AAAI}.

\bibitem[\protect\citename{Wei \bgroup et al.\egroup }2020]{wei2020CasRel}
Zhepei Wei, Jianlin Su, Yue Wang, Yuan Tian, and Yi~Chang.
\newblock 2020.
\newblock A novel cascade binary tagging framework for relational triple
  extraction.
\newblock In {\em Proceedings of ACL}.

\bibitem[\protect\citename{Yu and Lam}2010]{yu2010jointly}
Xiaofeng Yu and Wai Lam.
\newblock 2010.
\newblock Jointly identifying entities and extracting relations in encyclopedia
  text via a graphical model approach.
\newblock In {\em Proceedings of COLING}, pages 1399--1407.

\bibitem[\protect\citename{Yu \bgroup et al.\egroup }2020]{yu2020jointer}
Bowen Yu, Zhenyu Zhang, Xiaobo Shu, Yubin Wang, Tingwen Liu, Bin Wang, and
  Sujian Li.
\newblock 2020.
\newblock Joint extraction of entities and relations based on a novel
  decomposition strategy.
\newblock In {\em Proceedings of ECAI}.

\bibitem[\protect\citename{Zelenko \bgroup et al.\egroup
  }2003]{zelenko2003kernel}
Dmitry Zelenko, Chinatsu Aone, and Anthony Richardella.
\newblock 2003.
\newblock Kernel methods for relation extraction.
\newblock {\em Journal of machine learning research}.

\bibitem[\protect\citename{Zeng \bgroup et al.\egroup
  }2018]{zeng2018extracting}
Xiangrong Zeng, Daojian Zeng, Shizhu He, Kang Liu, Jun Zhao, et~al.
\newblock 2018.
\newblock Extracting relational facts by an end-to-end neural model with copy
  mechanism.
\newblock In {\em Proceedings of ACL}.

\bibitem[\protect\citename{Zeng \bgroup et al.\egroup }2019]{zeng2019learning}
Xiangrong Zeng, Shizhu He, Daojian Zeng, Kang Liu, Shengping Liu, and Jun Zhao.
\newblock 2019.
\newblock Learning the extraction order of multiple relational facts in a
  sentence with reinforcement learning.
\newblock In {\em Proceedings of EMNLP-IJCNLP}.

\bibitem[\protect\citename{Zhang \bgroup et al.\egroup
  }2019]{zhang2019bridging}
Wen Zhang, Yang Feng, Fandong Meng, Di~You, and Qun Liu.
\newblock 2019.
\newblock Bridging the gap between training and inference for neural machine
  translation.
\newblock In {\em Proceedings of ACL}.

\bibitem[\protect\citename{Zheng \bgroup et al.\egroup }2017]{zheng2017joint}
Suncong Zheng, Feng Wang, Hongyun Bao, Yuexing Hao, Peng Zhou, and Bo~Xu.
\newblock 2017.
\newblock Joint extraction of entities and relations based on a novel tagging
  scheme.
\newblock In {\em Proceedings of ACL}.

\end{thebibliography}

\end{document}